\newif\ifeusipstyle
\eusipstyletrue

\ifeusipstyle
  \documentclass[a4paper,9pt]{article}
  \usepackage{spconf,amsmath,graphicx,cite}
\else
  \documentclass[conference]{../../../tex/IEEEtran}
  \usepackage[dvips]{graphicx}
\fi

\usepackage{url}
\usepackage{amsfonts}

\title{A Neural Network Based on First Principles}

\ifeusipstyle
   \name{Paul M Baggenstoss}
   \address{Fraunhofer FKIE, Fraunhoferstr 20,
   \\ 53343 Wachtberg, Germany}
\else
   \author{\IEEEauthorblockN{Paul M. Baggenstoss}
   \IEEEauthorblockA{Fraunhofer FKIE,
   Fraunhoferstrasse 20\\
   53343 Wachtberg, Germany\\
   Email: p.m.baggenstoss@ieee.org}
   }
\fi

\begin{document}
\newcommand{\defined}{\stackrel{\mbox{\tiny$\Delta$}}{=}}
\newtheorem{example}{Example}
\newtheorem{conclusion}{Conclusion}
\newtheorem{assumption}{Assumption}
\newtheorem{definition}{Definition}
\newtheorem{problem}{Problem}
\newcommand{\erf}{{\rm erf}}

\newcommand{\sst}{\scriptstyle }
\newcommand{\xparen}{\mbox{\small$(\bfx)$}}
\newcommand{\hojz}{H_{0j}\mbox{\small$(\bfz)$}}
\newcommand{\Hozj}{H_{0,j}\mbox{\small$(\bfz_j)$}}
\newcommand{\smallmath}[1]{{\scriptstyle #1}}
\newcommand{\Hoz}[1]{H_0\mbox{\small$(#1)$}}
\newcommand{\Hozp}[1]{H_0^\prime\mbox{\small$(#1)$}}
\newcommand{\Hozpp}[1]{H_0^{\prime\prime}\mbox{\small$(#1)$}}
\newcommand{\hoz}{\Hoz{\bfz}}
\newcommand{\hooz}{\Hozp{\bfz}}
\newcommand{\hoooz}{\Hozpp{\bfz}}
\newcommand{\smJ}{{\scriptscriptstyle \! J}}
\newcommand{\smK}{{\scriptscriptstyle \! K}}

\newcommand{\erfc}{{\rm erfc}}
\newcommand{\bitem}{\begin{itemize}}
\newcommand{\dsum}{{ \displaystyle \sum}}
\newcommand{\eitem}{\end{itemize}}
\newcommand{\benum}{\begin{enumerate}}
\newcommand{\eenum}{\end{enumerate}}
\newcommand{\bdm}{\begin{displaymath}}
\newcommand{\bfzro}{{\underline{\bf 0}}}
\newcommand{\bfone}{{\underline{\bf 1}}}
\newcommand{\edm}{\end{displaymath}}
\newcommand{\beq}{\begin{equation}}
\newcommand{\bea}{\begin{eqnarray}}
\newcommand{\eea}{\end{eqnarray}}
\newcommand{\cali}{ {\cal \bf I}}
\newcommand{\caln}{ {\cal \bf N}}
\newcommand{\barray}{\begin{displaymath} \begin{array}{rcl}}
\newcommand{\earray}{\end{array}\end{displaymath}}
\newcommand{\eeq}{\end{equation}}
\newcommand{\btheta}{\mbox{\boldmath $\theta$}}
\newcommand{\bTheta}{\mbox{\boldmath $\Theta$}}
\newcommand{\blam}{\mbox{\boldmath $\Lambda$}}
\newcommand{\beps}{\mbox{\boldmath $\epsilon$}}
\newcommand{\bdelta}{\mbox{\boldmath $\delta$}}
\newcommand{\bgamma}{\mbox{\boldmath $\gamma$}}
\newcommand{\balpha}{\mbox{\boldmath $\alpha$}}
\newcommand{\bbeta}{\mbox{\boldmath $\beta$}}
\newcommand{\balphascript}{\mbox{\boldmath ${\scriptstyle \alpha}$}}
\newcommand{\bbetascript}{\mbox{\boldmath ${\scriptstyle \beta}$}}
\newcommand{\bLambda}{\mbox{\boldmath $\Lambda$}}
\newcommand{\bDelta}{\mbox{\boldmath $\Delta$}}
\newcommand{\bomega}{\mbox{\boldmath $\omega$}}
\newcommand{\bOmega}{\mbox{\boldmath $\Omega$}}
\newcommand{\blambda}{\mbox{\boldmath $\lambda$}}
\newcommand{\bphi}{\mbox{\boldmath $\phi$}}
\newcommand{\bpi}{\mbox{\boldmath $\pi$}}
\newcommand{\bnu}{\mbox{\boldmath $\nu$}}
\newcommand{\brho}{\mbox{\boldmath $\rho$}}
\newcommand{\bmu}{\mbox{\boldmath $\mu$}}
\newcommand{\sigi}{\mbox{\boldmath $\Sigma$}_i}
\newcommand{\bfu}{{\bf u}}
\newcommand{\bfx}{{\bf x}}
\newcommand{\bfb}{{\bf b}}
\newcommand{\bfk}{{\bf k}}
\newcommand{\bfc}{{\bf c}}
\newcommand{\bfv}{{\bf v}}
\newcommand{\bfn}{{\bf n}}
\newcommand{\bfK}{{\bf K}}
\newcommand{\bfh}{{\bf h}}
\newcommand{\bff}{{\bf f}}
\newcommand{\bfg}{{\bf g}}
\newcommand{\bfe}{{\bf e}}
\newcommand{\bfr}{{\bf r}}
\newcommand{\bfw}{{\bf w}}
\newcommand{\calX}{{\cal X}}
\newcommand{\calZ}{{\cal Z}}
\newcommand{\bb}{{\bf b}}
\newcommand{\bfy}{{\bf y}}
\newcommand{\bfz}{{\bf z}}
\newcommand{\bfs}{{\bf s}}
\newcommand{\bfa}{{\bf a}}
\newcommand{\bfA}{{\bf A}}
\newcommand{\bfB}{{\bf B}}
\newcommand{\bfV}{{\bf V}}
\newcommand{\bfZ}{{\bf Z}}
\newcommand{\bfH}{{\bf H}}
\newcommand{\bfX}{{\bf X}}
\newcommand{\bfR}{{\bf R}}
\newcommand{\bfF}{{\bf F}}
\newcommand{\bfS}{{\bf S}}
\newcommand{\bfC}{{\bf C}}
\newcommand{\bfI}{{\bf I}}
\newcommand{\bfO}{{\bf O}}
\newcommand{\bfU}{{\bf U}}
\newcommand{\bfD}{{\bf D}}
\newcommand{\bfY}{{\bf Y}}
\newcommand{\bSig}{{\bf \Sigma}}
\newcommand{\test}{\stackrel{<}{>}}
\newcommand{\zmk}{{\bf Z}_{m,k}}
\newcommand{\zlk}{{\bf Z}_{l,k}}
\newcommand{\zm}{{\bf Z}_{m}}
\newcommand{\ssq}{\sigma^{2}}
\newcommand{\dint}{{\displaystyle \int}}
\newtheorem{theorem}{Theorem}
\newcommand{\postscript}[2]{ \begin{center}
    \includegraphics*[width=3.5in,height=#1]{#2.eps}
    \end{center} }

\newtheorem{identity}{Identity}
\newtheorem{hypothesis}{Hypothesis}
\newcommand{\mathtiny}[1]{\mbox{\tiny$#1$}}

\maketitle

\begin{abstract}
In this paper, a Neural network is derived from first principles,
assuming only that each layer begins with a  linear dimension-reducing 
transformation. The approach appeals 
to the principle of Maximum Entropy (MaxEnt) 
to find the posterior distribution of the input data of each layer,
conditioned on the layer output variables.
This posterior has a well-defined mean, the conditional mean
estimator, that is calculated using a type of neural network
with theoretically-derived activation functions similar to sigmoid, softplus, and relu.
This implicitly provides a theoretical justification for their use.
A theorem that finds the conditional distribution and conditional mean estimator under the MaxEnt prior
is proposed, unifying results for special cases.  Combining 
layers results in an auto-encoder with conventional feed-forward
analysis network and a type of linear Bayesian belief network
in the reconstruction path.  
\end{abstract}

\section{Introduction}

\subsection{Motivation}
Despite the brilliant success of deep networks, networks and their activation functions are generally selected empirically
to learn general functions \cite{Goodfellow2016,NwankpaActFn2018}.  
In generative networks, the activation functions revolve around
approximating the expected value of generating 
distributions that are selected for tractability \cite{HintonRelu2010,zhou2016softplus,Ravanbakhsh} 
or are empirically determined \cite{ramachandran2017searching}.
Despite the elegant mathematical formulations, restricted Boltzmann machines (RBMs) \cite{WellingHinton04},
and variation autoencoders \cite{doersch2016tutorial}, the models 
are also selected based on tractability or empirical performance.
This paper seeks to derive the network structure and activation function from first principles
by deducing the network structure from the {\it a posteriori} distribution of the visible data given the layer output.

\subsection{Problem Statement}
\begin{figure}[h!]
  \begin{center}
    \includegraphics[width=3.5in]{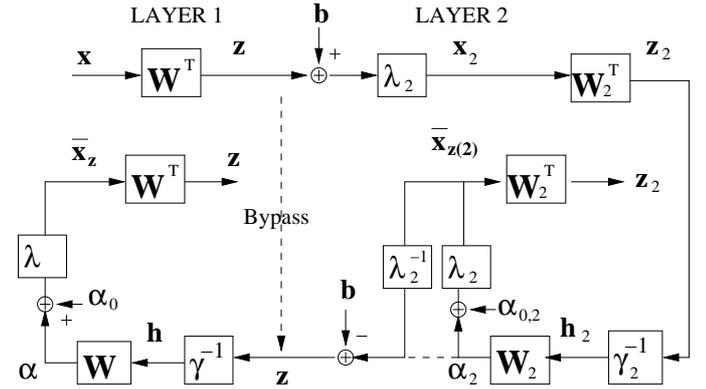}
  \caption{Block diagram.}
  \label{asy}
  \end{center}
  \vspace{-.2in}
\end{figure}

Figure \ref{asy} illustrates the main ideas of this paper.
The diagram shows two network layers, but we will focus on just the first layer for now.
The input to a network layer is a high-dimensional vector $\bfx\in \mathbb{R}^N$. A
lower-dimensional feature is computed by linear transformation,
$\bfz={\bf W}^\prime \bfx$, where $\bfz\in \mathbb{R}^M$ , and $M<N$.

A bias and activation function are applied prior to the next layer,
but this is not relevant to analyzing the first layer.
For now, the question is what can be inferred about 
$p(\bfx)$ from $\bfz$, bypassing layer 2 (See ``bypass" in Fig. \ref{asy}).
The remaining components in layer 1 are described below
and layer 2 is explained in Section \ref{seclyr}.

\section{Mathematical Approach}

\subsection{Prior Distribution}
The prior ({\it a priori} distribution) $p_0(\bfx)$
quantifies the expectation about $\bfx$ before feature $\bfz$ is measured.
The principle of maximum entropy (MaxEnt) \cite{Jaynes82} proposes
that the entropy of a distribution, given by
$
H\{p(\bfx)\}  = -\int_{\bfx} \; p(\bfx) \; \log(p(\bfx)) \; {\rm d} \bfx
$
should be as high as possible subject to the known constraints.
These distributions are generally of the exponential class \cite{Kapur}.
Consider the following univariate exponential class of distributions:
\vspace{-.1in}
\beq
  \log p(x;\alpha) =  \alpha x + b x^2  - \log Z(\alpha),
\vspace{-.1in}
  \label{ent1}
\eeq
where the dependence on $b$ has been removed from the notation because
it is fixed by the choice of prior. Parameter $\alpha$ plays a special role 
because it controls the distribution mean.
Let the expected value of distribution (\ref{ent1}) be written 
\vspace{-.1in}
\beq
  \lambda(\alpha) \defined \mathbb{E}\{ x;\alpha\}= \int_x \; x \; p(x;\alpha) \; {\rm d} x. 
\vspace{-.1in}
  \label{ldef1}
\eeq
In keeping with maximum entropy, $p_0(\bfx)$ should be constructed from
$N$ independent univariate distributions (\ref{ent1}) as follows
\beq
  \log p_0(\bfx) = \sum_{i=1}^N \; \log p(x_i;\alpha_0).
  \label{p0def}
\eeq
This class includes independent and identically distributed ({\it iid})
Gaussian, exponential, and their truncated variants, and they have highest entropy
among all multivariate distributions under constraints that will be proposed.

\subsection{Manifold Distribution}
Conditioned on knowing $\bfz$, $\bfx$ can only exist on the set
\beq
  {\cal M}(\bfz) = \{ \bfx : {\bf W}^\prime \bfx = \bfz\}.
\eeq
This is the set (a manifold) of all possible values of $\bfx$ that exactly
reproduce the measured value $\bfz$.
The posterior is therefore a manifold distribution 
\beq
 p(\bfx|\bfz) = p_0(\bfx) / \left( \int_{\bfx \in {\cal M}(\bfz)} \; p_0(\bfx) \; {\rm d} \bfx \right), \;\;\; \bfx \in {\cal M}(\bfz),
\label{pp0}
\eeq
which is $p_0(\bfx)$ projected onto the manifold, then normalized
so it integrates to 1.  To draw samples from (\ref{pp0}), samples 
are drawn from the  manifold ${\cal M}(\bfz)$ with probability proportional 
to the value of the prior distribution $p_0(\bfx)$.
It can be shown \cite{Bag_info} that the denominator in (\ref{pp0}) can be written
$$ \int_{\bfx \in {\cal M}(\bfz)} \; p_0(\bfx) \; {\rm d} \bfx = p_0(\bfz),$$
which is the prior feature distribution, i.e. distribution of $\bfz$ under the assumption that $\bfx \sim  p_0(\bfx)$.
Rewriting (\ref{pp0}),
\beq
 p(\bfx|\bfz) = \frac{p_0(\bfx)}{p_0(\bfz)}, \;\;\; \bfx \in {\cal M}(\bfz).
\label{pp1}
\eeq
Due to conditioning on $\bfz$, the denominator has a fixed value, so the manifold distribution is shaped
only by $p_0(\bfx)$.  This quantity is known in the method of PDF projection \cite{BagPDFProj, Bag_info}.
The conditional mean estimate is the mean of (\ref{pp0}), written
\vspace{-.1in}
\beq
\bar{\bfx}_z \defined {\displaystyle \int_{\bfx \in {\cal M}(\bfz)}  \bfx \; p(\bfx|\bfz)  {\rm d} \bfx}.
\label{xzbar}
\vspace{-.1in}
\eeq

\subsection{Main Result}
\label{thmsec}
Despite the simple form of (\ref{pp1}), it is not useful for sampling and (\ref{xzbar}) is not tractable.
Also, (\ref{pp1}) is not even a proper distribution, having infinite
density on an infinitely thin manifold.  To find a proper distribution that approximates (\ref{pp0}), 
we use a {\it surrogate density} \cite{BagUMS},  which  is a proper distribution that shares the properties of 
 (\ref{pp0}), which are (a) probability mass concentrated on the manifold ${\cal M}(\bfz)$, (b)
 mean $\bar{\bfx}_z \in {\cal M}(\bfz)$ (because ${\cal M}(\bfz)$ is convex), and (c) density on 
the manifold proportional to $p_0(\bfx)$.  The following theorem gives form to the surrogate density.
\vspace{0.1in}

\begin{theorem}
Let prior $p_0(\bfx)$ be written as (\ref{p0def}) with univariate densities $p(x;\alpha_0)$ of class (\ref{ent1})
with mean $\lambda(\alpha_0)$.   Then, the surrogate density for $p(\bfx|\bfz)$ can be written
\vspace{-.1in}
\beq
  \log p(\bfx;\balpha_0+\balpha) = \sum_{i=1}^N \; \log p(x_i;\alpha_0 + \alpha_i) ,
  \label{padef}
\vspace{-.1in}
\eeq
where $\balpha={\bf W} \bfh,$ and $\bfh$ is the solution of
\beq
{\bf W}^\prime \lambda\left(\balpha_0 + {\bf W} \bfh\right) = \bfz.
\label{tm1}
\eeq
Furthermore, the mean of $p(\bfx;\balpha_0+\balpha)$ equals
\vspace{-.1in}
\beq
  \hat{\bfx}_z = \lambda(\balpha_0+{\bf W} \bfh).
  \label{meanzh}
\eeq
\vspace{-.1in}
\label{thm1}
\end{theorem}

{\bf Outline of Proof:}. 
To show that solution $\bfh$ solving (\ref{tm1}) exists, it is shown below
that (\ref{tm1}) is the same as the saddle point (SP) equation
for the SP expansion of $p_0(\bfz)$.  Since for the exponential family (\ref{ent1}), the SP
expansion exists over the entire range of ${\bf z}$ (see \cite{barndorff1979edgeworth} appendix), 
the solution exists whenever $\bfz$ is valid, i.e. whenever 
$\bfz={\bf W}^\prime \bfx$ for a sample $\bfx$ in the support of $p_0(\bfx)$.
Since $\hat{\bfx}_z=\lambda\left(\balpha_0 + \balpha\right)$, 
${\bf W}^\prime \bar{\bfx}_z=\bfz,$ meeting property (b)  for a surrogate density.
Using (\ref{p0def}),(\ref{ent1}), 
the gradient of $\log p(\bfx;\balpha_0+\balpha)$ with respect to $\bfx$ is
$\left[\frac{\partial \log p(\bfx;\balpha_0+\balpha)}{\partial \bfx}\right] = \balpha_0 + \balpha + 2 b \bfx.$
In order that (\ref{padef}) is proportional to $p_0(\bfx)$ on the manifold,
it is necessary that the component of this gradient in any direction parallel
to the manifold (i.e. orthogonal to columns of ${\bf W}$) is the same
as for the prior $p_0(\bfx)$.  This can be mathematically written
${\bf B}^\prime \left[\balpha_0 + \balpha + 2 b \bfx\right] = {\bf B}^\prime \left[\balpha_0 + 2 b \bfx \right],$
for orthonormal matrix ${\bf B}$ spanning the linear subspace orthogonal to the columns of ${\bf W}$.
Therefore,  $\balpha$ must be fully orthogonal to ${\bf B}$, and of the form
$\balpha={\bf W}\bfh$. This fulfills property (c) of the surrogate density.
To fulfill property (a), it can be shown that the probability mass of the surrogate density
indeed converges to the manifold for large $N$ (see \cite{BagUMS}, Appendix A).
\vspace{.1in}\\

To see how Theorem \ref{thm1} defines a neural network, we first simplify notation, by defining
the function 
$\bfz = \gamma(\bfh) = {\bf W}^\prime \lambda \left( {\bf W} \bfh \right)$
and its inverse: $\bfh = \gamma^{-1}(\bfz).$  The concept of $\gamma^{-1}(\bfz)$ is illustrated in Figure \ref{asy}.
Feature $\bfz$, is converted to $\bfh$ through $\gamma^{-1}(\bfz)$, then multiplied by ${\bf W}$ to raise the dimension
back to $N$, and finally passed through activation function $\lambda(\;)$ to produce $\bar{\bfx}_z$.
Optionally, it can be passed to the generating distributions $p(\bfx; \balpha_0+\balpha)$ for stochastic generation.
According to the definition of $\gamma^{-1}(\;)$,  ${\bf W}^\prime \bar{\bfx}_z = \bfz,$
or in other words, the feature $\bfz$ is recovered exactly when  $\bar{\bfx}_z$ is processed
by the forward path, illustrated in the figure where $\bfz$ of the forward path is identified with  $\bfz$ produced by the circular path.
In this role, $\gamma^{-1}(\bfz)$ acts as a non-linearity (but is not applied element-wise). 
Despite the iterative solution of $\gamma^{-1}(\bfz)$, its derivatives
are easly calculated from $\gamma(\bfh)$, so are amenable to back-propagation training
for optmizing the network parameters.

Since $\hat{\bfx}_z$ is the conditional mean estimator, it
enjoys numerous optimal properties such as minimum mean square estimator \cite{KayEst}.  
Since the surrogate density converges to the posterior $p(\bfx|\bfz)$,
it implies that $\hat{\bfx}_z \rightarrow \bar{\bfx}_z$ for large $N$.
This convergence occurs quickly and low dimension as has been demonstrated 
in certain cases (see fig. 8 in \cite{BagUMS}).  
In fact, $\hat{\bfx}_z=\bar{\bfx}_z$ under certain symmetry conditions.
A special case of (\ref{meanzh})  corresponds to autoregressive spectral estimation,
which can be generalized for conditioning on any linear function of the spectrum, such as
MaxEnt inversion of MEL band features \cite{BagUMS}.  Another 
special case of (\ref{meanzh}) is mathematically the same as
classical maximum entropy image reconstruction \cite{Wernecke77,Wei87}.
It is also not surprising, given form (\ref{pp1}), that the
surrogate density has a close relationship to $p_0(\bfz)$.
It can be shown that $\bfh$ is also the saddlepoint for the SP approximation to $p_0(\bfz)$.
This can be seen by comparing  (\ref{tm1}) with equation 25 in \cite{BagNutKay2000}, page 2245,
which is the general SP equation for the linear sum of independent random variables.
It can also be shown that $\bfh$ is the maximum likelihood estimate  under the 
likelihood function (\ref{padef}) \cite{barndorff1979edgeworth}.

\section{Three cases of $\mathbb{X}$}
The MaxEnt prior $p_0(\bfx)$ depends on the range of $\bfx$, denoted by $\mathbb{X}$,
and any other assumptions.  The choice of $p_0(\bfx)$ in turn determines the activation function $\lambda(\alpha)$
and how to sample from the manifold ${\cal M}(\bfz)$.
Whereas manifold sampling is exact sampling of $p(\bfx|\bfz)$,
sampling from the surrogate density is an approximation. 
However, experiments have demonstrated the almost perfect correspondence
between the two distributions (e.g. Figures 8,10,11 in \cite{BagPBN}).

\subsection{ Unit hypercube $\mathbb{U}^N$}
In the {\it unit hypercube}, denoted by $\mathbb{U}^N$,  elements of $\bfx$ are in the range $[0, \; 1]$,
the case for intensity images, or if  $\bfx$ is the output of a sigmoid activation function.
 The uniform prior is the MaxEnt distribution in $[0,\;1]$,  $p_0(\bfx) = 1$, which is the trivial case of (\ref{ent1})
with  $\alpha_0=0$, $b=0$.  Sampling from  ${\cal M}(\bfz)$ uniformly within $\mathbb{U}^N$ is done
 using a type of Monte Carlo Markov chain (MCMC) called hit-and-run \cite{SmithUniform}, with modification
for $\mathbb{U}^N$ as explained in detail in (\cite{BagUMS}, Sec. V, p. 2465).
The surrogate density is a truncated exponential distribution (TED)
 $p(x; \alpha) = \frac{\alpha}{e^\alpha-1} \; e^{\alpha x}, \;\; 0>x>1$. The activation function is 
the TED nonlinearity \cite{BagEusipcoRBM,BagUMS} 
\vspace{-.1in}
\beq
  \lambda(\alpha)=\frac{e^\alpha}{e^\alpha - 1} - \frac{1}{\alpha}
  \label{teddef}
\eeq
which resembles the sigmoid (see Fig. \ref{ted-tg}).  This problem has been studied in detail in (\cite{BagUMS}, Sec. V, p. 2465).

\subsection{ Positive Quadrant $\mathbb{P}^N$}
    We assume that elements of $\bfx$ are positive, so exist
     in the positive quadrant of $\mathbb{R}^N$, denoted by $\mathbb{P}^N$. This happens if
      $\bfx$ is the output of an previous network layer and a rectifying activation
   function was used, or if $\bfx$ is some kind of spectral or intensity data that is inherently positive.
      There is no proper MaxEnt distribution on the open interval $[0, \; \infty]$ without
      constraining the mean or variance, resulting in two solutions.
      The constrained mean case results in an exponential prior  and has been
     studied in detail (\cite{BagUMS}, Sec. IV, p. 2460). Although it corresponds to
     MaxEnt image reconstruction \cite{Wernecke77,Wei87}, it  is less interesting for
     neural networks.

     If we are willing to assume a fixed variance, the
    truncated Gaussian with mean parameter 0 and variance parameter 1 (not the same
    as mean 0 and variance 1) provides the distribution with maximum entropy on $[0, \; \infty]$ \cite{Kapur}.
      This is the case of (\ref{ent1}) with  $\alpha_0=0$, $b=1$.   
      This can also be written 
     \vspace{-.2cm}
     \beq
      p_0(\bfx) = \prod_{i=1}^N \; 2 {\cal N}(x_i), \;\; \; x_i>0, \forall i,
    \label{tg0}
     \vspace{-.2cm}
     \eeq
     where 
     $ {\cal N}(x) = \frac{e^{-x^2/2}}{\sqrt{2\pi}}.$
     To sample ${\cal M}(\bfz)$ with this prior, an MCMC method similar to the exponential case
     (given in (\cite{BagUMS}, Sec. IV, p. 2460) can be used.  
  A program implementing this procedure is provided in the appendix. 
  The activation function is the mean of the truncated Gaussian:
     \vspace{-.1cm}
  \beq
 \lambda(\alpha) =  \alpha + \frac{{\cal N}(\alpha)}{\Phi(\alpha)}
     \vspace{-.1cm}
   \label{tgdef}
 \eeq
   which resembles softplus (see Figure \ref{ted-tg}).

\subsection{ Unconstrained $\mathbb{R}^N$}
   \label{gaussec}
      There is no proper MaxEnt distribution on the open interval $[-\infty, \; \infty]$ without
    constraining the variance.  In many cases, data has been normalized, so we are justified in using
 a standard Gaussian prior, which is the MaxEnt distribution on $\mathbb{R}^N$ for known variance \cite{Kapur}.
For this case, the surrogate density is the same as the exact posterior.
All samples on the manifold ${\cal M}(\bfz)$  can be written 
$\bfx = \bar{\bfx}_z +  {\bf B} \bfu,$ where $\bar{\bfx}_z= {\bf W} \bfh$.  
and ${\bf B}$ is the same as in the proof of Theorem \ref{thm1}.
This case is particularly instructive because $\gamma^{-1}(\bfz)$ has the closed-form
expression $\gamma^{-1}(\bfz)=\left( {\bf W}^\prime {\bf W}\right)^{-1} \bfz,$
so applying $\gamma^{-1}(\bfz)$ corresponds to least-squares.
To conform to the assumed prior distribution,
$\bfu$ is a set of $(N-M)$ independent Gaussian random variables of zero mean and
variance 1.  The activation function is linear, $\lambda(x)=x$. 

\subsection{Summary and Remarks}
\begin{figure}[h!]
  \begin{center}
    \includegraphics[width=3.5in,height=1.2in]{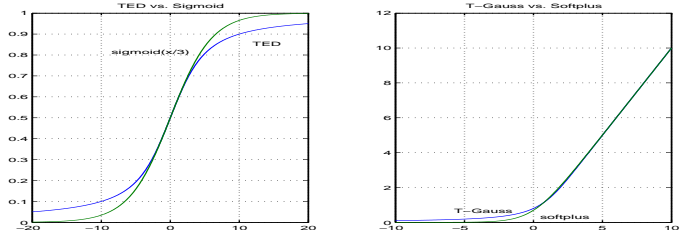}
  \caption{Left: TED activation compared to Sigmoid. Right: TG activation compared to Softplus.}
  \label{ted-tg}
  \end{center}
\end{figure}

Above results are summarized in the following table 
for four combinations of $\mathbb{X}$, 
and constraints (Const.).  For each case, the table provides
the MaxEnt prior $p_0(\bfx)$, the univariate distribution
 $p(x;\alpha)$ used to form the surrogate posterior (\ref{padef}),
and the mean function $\lambda(\alpha)$.
%
\begin{center}
 \begin{tabular}{|l|l|l|l|l|}
\hline
$\mathbb{X}$           & Const.            & $p_0(\bfx)$  & $p(x;\alpha)$                                 & $\lambda(\alpha)$  \\
\hline
 $\mathbb{U}^N$ &  N/A               & 1            &  $\frac{\alpha}{e^\alpha-1} \; e^{\alpha x}$  &  $\frac{e^\alpha}{e^\alpha - 1} - \frac{1}{\alpha}$ \\
\hline
 $\mathbb{P}^N$ & $ \mathbb{E}(x)$   &  $e^{-x}$  & $\alpha e^{-\alpha x}$ & ${1 \over \alpha}$ \\
\hline
 $\mathbb{P}^N$ & $\mathbb{E}(x^2)$  &   $2 {\cal N}\left(x\right)$  & $\frac{ {\cal N}\left(x-\alpha\right)}{ \Phi\left( \alpha\right)}$ & $\alpha + \frac{{\cal N}(\alpha)}{\Phi(\alpha)}$ \\
\hline
$\mathbb{R}^N$ & $\mathbb{E}(x^2)$    &   ${\cal N}\left(x\right)$  & ${\cal N}\left(x-\alpha\right)$ & $\alpha$ \\
\hline
\end{tabular}
\\
where ${\cal N}\left(x\right) \defined \frac{e^{-x^2/2}}{\sqrt{2\pi}}$ and $\Phi\left( x\right)  \defined \int_{-\infty}^x {\cal N}\left(x\right) .$
\end{center}
The functions $\lambda(\alpha)$ are the MaxEnt ``activation functions" and
resemble commonly-used functions (see Fig. \ref{ted-tg}).
Note that the truncated Gaussian (TG) nonlinearity approaches the rectified linear unit (RELU) as the assumed variance of the prior
(normally equal to 1) goes to zero.
Alternatively, $\lambda(\;)$ can be replaced by the
generating distribution $p(\bfx; \balpha_0+\balpha)$
for stochastic generation.  For a single layer, this would
produce an RBM with deterministic forward path.

\section{Building a Network}
\label{seclyr}
In Figure \ref{asy}, a 2-layer network is created by
adding another MaxEnt layer.
The forward path (top) is a standard feed-forward network employing the
MaxEnt activation functions.  The data is first passed through a bias and activation function
$\lambda_2(\;)$ before being presented to the second layer's
linear transform. Note that after layer 2 reconstructs its input
in the backward path ($\bar{\bfx}_{z(2)}$) the activation function $\lambda_2(\;)$ and bias must be 
inverted before being processed by $\gamma^{-1}$.  However, because the forward activation function $\lambda_2(\;)$
is the same as the MaxEnt activation function for layer 2, then $\lambda_2(\;)$
cancels $\lambda_2^{-1}(\;)$, resulting in a simplified
backward path (see short dotted line at the bottom of the figure).  
It is also worth noting that in the backward
(reconstruction) path, stochastic generation using $p(\bfx; \balpha_0+\balpha)$ can be used
in place of activation functions to create stochastic networks.

The reverse path (bottom) consists of applying $\gamma^{-1}(\bfz)$
(after removal of bias, if needed), followed by dimension-increasing
transformation by the layer weight matrices (same matrix used in the
forward path). This eliminates the need for separate reconstruction 
weights, and decreases network parameter count.
This has been called a deterministic projected
belief network \cite{BagPBN,BagEusipcoPBN}.

Although the existence of $\bfh=\gamma^{-1}(\bfz)$ is guaranteed for a single layer,
it is not guaranteed for multiple layers. In other words, if $\bfz$ applied to 
$\gamma^{-1}$ is derived from the second layer and not from the forward path of the first layer,
then $\bfh=\gamma^{-1}(\bfz)$ is not guaranteed to exist.
This is the {\it sampling efficiency} issue on projected belief networks \cite{BagEusipcoRBM}.
It has been experimentally shown that as a PBN is trained, the sampling efficiency approaches 1.0  \cite{BagEusipcoRBM}.

\section{Conclusions}
In this paper, a new theorem has been presented that 
provides a closed-form asymptotic (large $N$) expression for the conditional mean $\bar{\bfx}_z = \mathbb{E}\{\bfx | \bfz\}$
given the output $\bfz$ of a dimension-reducing linear transformation under a class of
MaxEnt prior distributions.  The computation of the conditional mean resembles a linear Bayesian belief network layer
with special non-linear function preceding the linear transformation and special
activation function.  
Methods to sample the posterior $p(\bfx| \bfz)$ are provided.  
Applying this concept results in an auto-endoding
neural network based on first principles.  

\appendix[ Appendix: {\bf Sampling from ${\cal M}(\bfz)$ under the truncated Gaussian Prior}]
\label{app1}
Let ${\bf B}$ be the same as in the proof of Theorem \ref{thm1},
so is a $N \times (N-M)$ orthonormal matrix orthogonal to ${\bf W}$.
Let $\bfx$ be an $N\times 1$ vector, that is a member of ${\cal M}(\bfz)$, 
so ${\bf W}^\prime \bfx = \bfz.$
The following procedure will find a new candidate $\bfx$, and if used repeatedly, will generate
samples distributed according to (\ref{pp1}) with $p_0(\bfx)$ as given by
(\ref{tg0}). The following MATLAB program will generate
samples in $\mathbb{P}^N$ and in ${\cal M}(\bfz)$ uniformly or 
according to the truncated Gaussian prior (see comments). 
{\footnotesize
\begin{verbatim}
function x=ums_tgauss_iter(x,B);
   [n,bdim]=size(B);
   bu=zeros(bdim,1);  
   if(any(x<=0)), error('invalid data') end;
   for j=1:bdim,
      xu = B(:,j)./x;
      maxuneg=-1/max(xu);
      maxupos=-1/min(xu);
      gap  = (maxupos-maxuneg);

      %    % uniform sampling
      %    newduj = maxuneg + gap*rand;
      %    x = x + B(:,j) * newduj;

      % Trunc-Gauss sampling
      bn=norm(B(:,j));
      Bn=B(:,j)/bn;
      xb = Bn'*x;
      lims=sort([ xb+bn*maxuneg, xb+bn*maxupos]);
      bu(j) = sample_tgauss(lims(1),lims(2));
      x = x + B(:,j) * (bu(j)-xb);
  end;
return
% Note: function  sample_tgauss(a,b)  samples a 
% standard Truncated Gaussian in range [a,b], 
\end{verbatim}
}

\bibliographystyle{ieeetr}
\bibliography{ppt}
\end{document}